\pdfoutput=1

\documentclass[11pt]{article}

\usepackage[dvipsnames,svgnames,table]{xcolor}
\makeatletter

\usepackage[preprint]{acl}

\usepackage{times}
\usepackage{latexsym}
\usepackage[T1]{fontenc}
\usepackage[utf8]{inputenc}
\usepackage{microtype}
\usepackage{inconsolata}
\usepackage{graphicx}
\usepackage{booktabs}
\usepackage{xspace}
\usepackage[most]{tcolorbox}
\newcommand{\inftybench}[0]{$\infty$Bench\xspace}
\newcommand{\quality}[0]{QuALITY\xspace}
\newcommand{\narrativeqa}[0]{NarrativeQA\xspace}
\newcommand{\dosrag}[0]{DOS~RAG\xspace}
\newcommand{\vanillarag}[0]{Vanilla~RAG\xspace}
\newcommand{\gptFourO}[0]{GPT-4o\xspace}
\newcommand{\gptFourOmini}[0]{GPT-4o-mini\xspace}

\newtcolorbox[list inside=prompt,auto counter,number within=section]{prompt}[1][]{
    colbacktitle=black!60,
    coltitle=white,
    fontupper=\footnotesize,
    boxsep=5pt,
    left=0pt,
    right=0pt,
    top=0pt,
    bottom=0pt,
    boxrule=1pt,
    #1,
}

\title{Stronger Baselines for Retrieval-Augmented Generation\\with Long-Context Language Models}

\author{Alex Laitenberger \and Christopher D. Manning \and Nelson F. Liu \\
       Stanford University, USA \\       \texttt{alex.laitenberger@gmail.com / alaiten@stanford.edu,}\\  \texttt{\{manning, nfliu\}@cs.stanford.edu} \\}

\begin{document}
\maketitle
\begin{abstract}
With the rise of long-context language models (LMs) capable of processing tens of thousands of tokens in a single context window, do multi-stage retrieval-augmented generation (RAG) pipelines still offer measurable benefits over simpler, single-stage approaches? 
To assess this question, we conduct a controlled evaluation for QA tasks under systematically scaled token budgets, comparing two recent multi-stage pipelines, ReadAgent and RAPTOR, against three baselines, including \dosrag (\underline{D}ocument's \underline{O}riginal \underline{S}tructure \underline{RAG}), a simple retrieve-then-read method that preserves original passage order.  
Despite its straightforward design, \dosrag consistently matches or outperforms more intricate methods on multiple long-context QA benchmarks. We trace this strength to a combination of maintaining source fidelity and document structure, prioritizing recall within effective context windows, and favoring simplicity over added pipeline complexity.
We recommend establishing \dosrag as a simple yet strong baseline for future RAG evaluations, paired with state-of-the-art embedding and language models, and benchmarked under matched token budgets, to ensure that added pipeline complexity is justified by clear performance gains as models continue to improve.\footnote{We release our code at \url{https://github.com/alex-laitenberger/stronger-baselines-rag}.}
\end{abstract}

\begin{figure}[t]
  \includegraphics[width=\columnwidth]{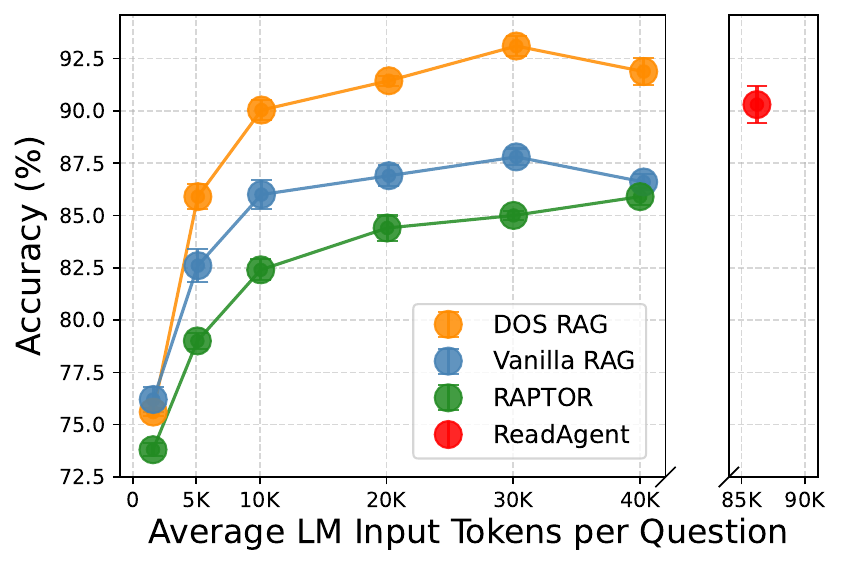}
  \caption{\inftybench En.MC performance of various multi-stage RAG systems and long-context baselines (mean $\pm$ standard deviation over five runs). All methods use \gptFourO as the underlying reader.
  For token budgets greater than 5K, \dosrag outperforms the complex multi-stage methods (ReadAgent and RAPTOR) by 2--8 points.}
  \label{fig:plot-infinity-4o}
\end{figure}

\section{Introduction}

Recent advances in long-context language models (LMs) have expanded their token processing capabilities, enabling them to handle tens of thousands of tokens in a single context window. This raises a pivotal question: Are complex, multi-stage retrieval-augmented generation (RAG) pipelines still necessary when simpler, single-stage methods can now leverage these extended contexts effectively?

RAG systems traditionally combine a \emph{retriever}, which selects passages from a large corpus relevant to a given query, and a \emph{reader}, typically an LM, to generate a final answer \citep{lewis2021retrievalaugmentedgenerationknowledgeintensivenlp}. 
de Prior work has proposed a variety of complex, multi-stage retrieval strategies to circumvent the limited long-context reasoning ability of earlier reader LMs. For example, abstractive preprocessing, iterative passage summarization, and agent-based retrieval loops have been used to compress or reason over documents that might otherwise exceed the input limits of early LMs \citep[\emph{inter alia}]{chen2023walkingmemorymazecontext,sarthi2024raptorrecursiveabstractiveprocessing,lee2024humaninspiredreadingagentgist,sun-etal-2024-pearl}. While effective, these pipelines often introduce significant complexity and computational overhead.

\begin{figure*}[t]
  \includegraphics{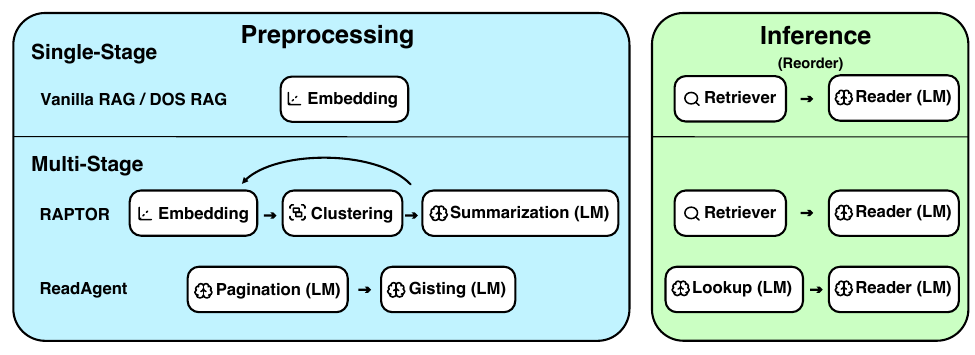}
  \caption{Comparison of single-stage vs.\ multi-stage RAG pipelines. Vanilla RAG/\dosrag use a minimal retrieve-then-read setup, while RAPTOR and ReadAgent add additional preprocessing and LM-based steps (e.g., clustering, iterative summarization, pagination, gisting, lookup), increasing pipeline complexity and cost.
  }
  \label{fig:rag-methods}
\end{figure*}

In contrast, modern long-context LMs can now directly process substantial amounts of text, suggesting that simpler retrieve-then-read strategies might suffice in certain settings.  
To compare multi-stage pipelines vs.\ simpler retrieve-then-read strategies, we conduct a controlled evaluation in which we systematically increase token budgets, analyzing how effectively each approach leverages extended contexts using a representative modern long-context LM (\gptFourO) as the downstream reader (see \S\ref{sec:exp_setup}).
We compare two recent multi-stage pipelines (ReadAgent and RAPTOR; \citealp{lee2024humaninspiredreadingagentgist,sarthi2024raptorrecursiveabstractiveprocessing}) against three baselines, including \dosrag (\underline{D}ocument's \underline{O}riginal \underline{S}tructure \underline{RAG}).  
\dosrag maintains a simple retrieve-then-read strategy and presents retrieved passages in their original document order. Despite its simplicity, our findings across three QA benchmarks (\inftybench, \quality, \narrativeqa) indicate that \dosrag can match or even outperform more complex multi-stage pipelines on all evaluated retrieval token budgets (see \S\ref{sec:results}). Our analysis suggests that \dosrag's strength lies in preserving original passages and document structure, prioritizing recall within effective context windows, and maintaining simplicity over added pipeline complexity (see \S\ref{sec:analysis}). 

This work advocates for establishing \dosrag as a simple yet strong baseline for RAG evaluations, paired with state-of-the-art embedding and language models and benchmarked under matched token budgets, so that added pipeline complexity is justified only when it delivers clear performance gains as model capabilities continue to evolve.

\section{Experimental Setup}
\label{sec:exp_setup}

We compare the performance of two recent multi-stage RAG pipelines (ReadAgent and RAPTOR) against three baselines (\vanillarag, the full-document baseline, and \dosrag) on three long-context question-answering tasks (\inftybench, \quality and \narrativeqa).
See Figure~\ref{fig:rag-methods} for a visual method overview and Appendix~\ref{app:experimental-setup-details} for further details about experimental setup, implementation, and used prompts.

\subsection{Benchmarks}

\paragraph{\inftybench.}
We evaluate systems on the English multiple-choice (En.MC) subset of \inftybench \citep{zhang2024inftybenchextendinglongcontext}. The benchmark contains 229 multiple-choice questions on 58 documents (average length of 184K tokens).

\paragraph{\quality.}
We use the \quality benchmark \citep{pang2022qualityquestionansweringlong}, a multiple-choice question-answering dataset over English context passages containing between 2K to 8K tokens.
We evaluate systems on the development set, which contains 115 documents and 2,086 questions.

\paragraph{\narrativeqa.}
The \narrativeqa benchmark is a long-document question-answering dataset in English that challenges models to answer questions about stories by reading entire books or movie scripts \citep{kocisky-etal-2018-narrativeqa}. We evaluate on the test set, which contains 355 stories (avg. 57K tokens, up to 404K) and 10,557 questions. Each story's questions are constructed such that high performance requires understanding the underlying narrative, versus relying on shallow pattern matching.

\subsection{Multi-Stage RAG Pipelines}

\begin{figure}[t]
  \includegraphics[width=\columnwidth]{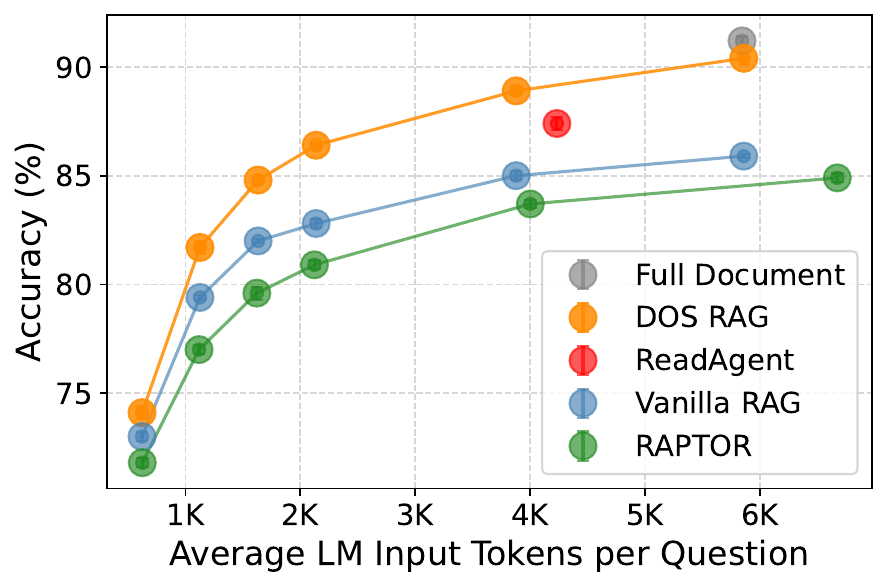}
  \caption{QuALITY performance of various multi-stage RAG systems and long-context baselines. All methods use \gptFourO as the underlying reader.
  Prompting long-context language models with entire documents (the full-document baseline) outperforms retrieval-augmented approaches, while \dosrag performs the best under token budget constraints.}
  \label{fig:quality-4o-plot}
\end{figure}

\paragraph{ReadAgent.} ReadAgent handles long input contexts with a method inspired by human reading strategies \citep{lee2024humaninspiredreadingagentgist}.
Concretely, ReadAgent prompts the LM with three steps: (1)~episode pagination, where the LM forms a sequence of pages by identifying natural breakpoints in the text; (2)~memory gisting, which compresses the content of each page into shorter ``gist'' summaries; and (3)~interactive look-up, where the LM uses the query and the gists to identify pages to re-read and use to solve the final query.
This approach extends the language model's context window by offloading the document's full detail into a page-wise gisted memory, retrieving original text only when needed. 

\paragraph{RAPTOR.}

RAPTOR handles long documents by recursively organizing the text into a tree of hierarchical summaries \citep{sarthi2024raptorrecursiveabstractiveprocessing}.
Concretely, it partitions the text into sentence-level passages, clusters related passages, and uses a language model to summarize each cluster.
This process repeats, generating higher-level summaries until a final set of root nodes represents the entire document.
At inference time, RAPTOR retrieves from different levels of the summary tree, balancing broad coverage against local detail.

\subsection{Baselines}

Our three baselines are designed to benefit from and scale with stronger language models with improved long-context reasoning abilities.

\paragraph{\vanillarag.}

In our implementation the document is first split into passages capped at 100 tokens, while preserving sentence boundaries where possible. We use neural retrieval with a sentence embedding model (Snowflake Arctic-embed-m 1.5 by \citealp{merrick2024embeddingclusteringdataimprove}) to encode both the query and the resulting passages into a shared embedding space. At inference time, passages are ranked by cosine similarity to the query embedding, with the top-ranked passages retrieved until a fixed input token budget (e.g., 10K tokens) is reached. The selected passages, ordered by decreasing similarity, are then concatenated with the query to construct the input to the language model.

\paragraph{Full-Document Baseline.}

Standard RAG pipelines do not preserve the narrative structure within documents, as passages are concatenated solely by retrieval rank.  
Moreover, retrieval errors can propagate to the downstream language model, which must then reason over potentially missing and disjoint content.  
To better understand how long-context LMs handle such challenges, we compare against a full-document baseline that simply prompts the model with \emph{all} available text---eliminating the need to filter passages. We evaluate this baseline only on the \quality benchmark, where all documents fit within the language model's context window.

\paragraph{Using the \underline{D}ocument's \underline{O}riginal \underline{S}tructure (\dosrag).}

\dosrag follows the same retrieval and embedding process as \vanillarag, but with one key difference: retrieved passages are reordered to match their original order in the document, not sorted by similarity score. This reordering, achieved by tracking passage positions, preserves original passage order like the full-document baseline while still filtering irrelevant content like \vanillarag. 

Formally, given a query $q$ and a document $d$ segmented into passages $(p_1, p_2, \ldots, p_n)$, let $\operatorname{sim}(q, p)$ denote the similarity score between $q$ and passage $p$.
Vanilla RAG retrieves and orders a subset of passages as 
\[(p_{i_1}, p_{i_2}, \ldots, p_{i_k})
\quad \text{where} \]
\[ \quad
\mathrm{sim}(q, p_{i_1}) \geq \mathrm{sim}(q, p_{i_2}) \geq \cdots \geq \mathrm{sim}(q, p_{i_k}).
\]
In contrast, \dosrag reorders the same retrieved passages by their original position in the document as
\[
(p_{j_1}, p_{j_2}, \ldots, p_{j_k})
\quad \text{where} \quad
j_1 < j_2 < \cdots < j_k.
\]

\section{Results}
\label{sec:results}

\begin{figure}[t]
  \includegraphics[width=\columnwidth]{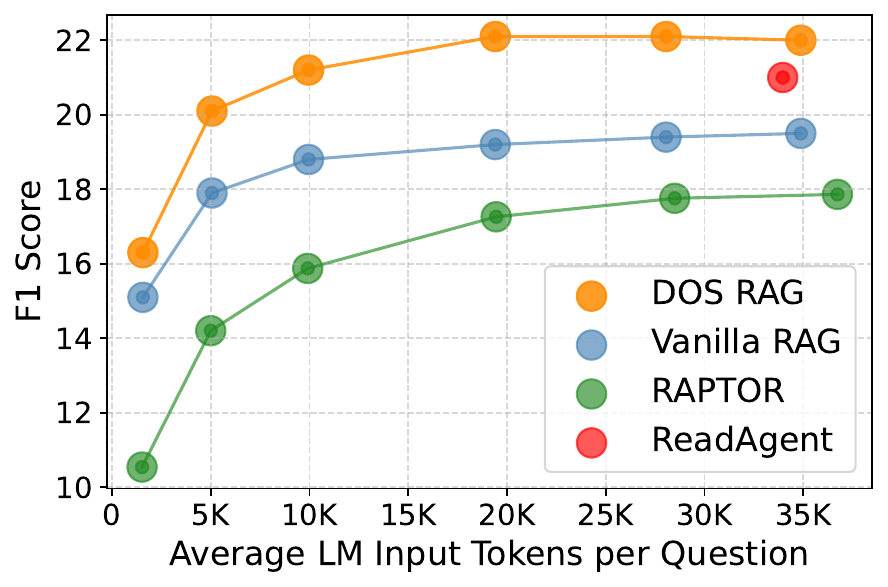}
  \caption{NarrativeQA performance of various multi-stage RAG systems and long-context baselines. All methods use \gptFourOmini as the underlying reader.
  At each evaluated token budget, \dosrag outperforms multi-stage retrieval systems and \vanillarag.}
  \label{fig:narrativeqa-4o-mini-plot}
\end{figure}

On all of \inftybench, \quality, and \narrativeqa we find that \dosrag performance consistently surpasses or matches complex multi-stage systems. See Appendix~\ref{app:full_results} for full results tables for all evaluated methods and benchmarks.

\paragraph{\inftybench.} 
Figure~\ref{fig:plot-infinity-4o} summarizes performance under varying retrieval token budgets (from 1.5K to 40K tokens) when using \gptFourO as the reader.

At 30K tokens, \dosrag achieves 93.1\%, outperforming \vanillarag (87.8\%) and both multi-stage methods by 2--8 points. Despite consuming more tokens (86K on average), ReadAgent underperforms \dosrag at moderate budgets (20K), highlighting the diminishing returns of multi-stage complexity when a single-pass prompt can already incorporate the relevant context.
 
Finally, we see that \dosrag performance begins to plateau as the retrieval budget grows beyond 30K tokens, while \vanillarag and RAPTOR also saturate at lower accuracies.

\paragraph{\quality.} Figure~\ref{fig:quality-4o-plot} shows performance on the \quality benchmark, again with \gptFourO as the reader model. In this setting, we see that all approaches show a steady rise in accuracy as the retrieval budget grows.
In particular, \emph{full-document baseline} with \gptFourO achieves 91.2\%, outperforming the best retrieval-augmented systems.
Among the retrieval-augmented methods, \dosrag again achieves the highest performance for token budgets of up to 8K.

\paragraph{\narrativeqa.} Figure~\ref{fig:narrativeqa-4o-mini-plot} presents the results for \narrativeqa across retrieval token budgets ranging from 1.5K to 40K, with \gptFourOmini as the reader.
Once again, we find that ReadAgent and RAPTOR consistently underperform \dosrag.
In particular, \dosrag achieves superior results while using only one third of the tokens required by ReadAgent.
These trends remain consistent across five different evaluation metrics (see Table~\ref{tab:narrative-test-4o-mini} in Appendix~\ref{app:narrativeqa_results} for detailed results).

\section{Analysis}
\label{sec:analysis}

\paragraph{Why is \dosrag effective?}

We identify four key factors that underlie \dosrag's performance and are supported by empirical findings from our evaluation:

1. Retrieving from \emph{original passages} rather than \emph{generated summaries}, thereby preserving source information, as in \vanillarag and the full-document baseline: We implemented \vanillarag as an exact ablation of RAPTOR that excludes generated summaries from the retrieval process. \vanillarag consistently outperforms RAPTOR across datasets and retrieval sizes, reinforcing our hypothesis that retrieving directly from original passages results in more robust QA, particularly as long-context LMs reduce the need for intermediate abstraction. While RAPTOR demonstrated superiority in the original paper, its key ablation used UnifiedQA-3B \citep{khashabi-etal-2020-unifiedqa} as the reader model, restricted to a 400-token input. Such a setting highlights RAPTOR's benefits under constrained context and model capacity, but does not generalize to today's stronger long-context LMs. Our results with \gptFourO show that, once stronger models and larger context windows are available, retrieving directly from original passages tends to be more robust. This illustrates how advances in model capacity and context length can shift the relative effectiveness of different pipeline designs.

2. Prioritizing \emph{retrieval recall over precision} while staying within the LM's \emph{effective context size}: \dosrag's performance increases consistently as the retrieval budget expands up to 30K tokens, after which it plateaus and declines, aligning with prior findings that LMs' effective context length remains limited (\citealp{liu-etal-2024-lost}). For shorter documents (6K--8K tokens), the full-document baseline outperforms all methods, indicating that maximizing recall, by including critical information \emph{anywhere} in the input, can be more effective than precision filtering. However, beyond the effective context window, eliminating irrelevant passages remains essential to maintain performance.

3. \emph{Reordering retrieved passages} to maintain narrative and argument continuity: \vanillarag serves as an exact ablation of \dosrag, excluding the reordering step. Across all benchmarks and retrieval budgets, \dosrag consistently outperforms \vanillarag, underscoring the benefits of preserving passage order. Performance gain is especially high when the retrieval budget is expanded to tens of thousands of tokens. Retrieving more passages brings us closer to the original document, but without order, the input becomes a disjointed, shuffled version.

4. Favoring \emph{simple over complex} pipelines: Multi-stage, agentic approaches like ReadAgent decompose QA into multiple LM calls, increasing token usage and latency. However, our evaluation shows that this added complexity does not necessarily improve performance. ReadAgent underperforms compared to \dosrag at lower token budgets, highlighting the effectiveness of simpler RAG pipelines that use strong embedding models and long-context LMs.

\section{Related Work}

Our results contribute to a growing body of work on comparing and combining retrieval-augmented methods against and with long-context LMs.

In particular, a variety of past work has studied whether retrieval remains necessary in the retrieve-then-read setting as language models gain better long-context reasoning capabilities. However, conclusions differ over time depending on the long-context abilities of the specific LMs used in experiments.
For example, \citet{xu2024retrieval} show that a 4K-context LM (Llama2-70B) with simple retrieval augmentation matches the performance of a context-extended 16K-context Llama2-70B model prompted with the full document, while using far less computation.
\citet{li-etal-2024-retrieval} revisit this question with a stronger long-context language model (GPT-4, with 32K token context) and find that directly prompting it with entire documents outperforms retrieval-augmented methods on several benchmarks, but at the cost of requiring substantially higher input token budgets.
Finally, work by  \citet{yu2024defenserageralongcontext} shows that preserving the original document order when prompting (i.e., as done in our \dosrag baseline) improves retrieval-augmented performance beyond the long-context full-document baseline.

In contrast, rather than debating the merits of retrieval vs.\ long-context language models, our work compares the \emph{combination} of retrieval and long-context language models (e.g., \dosrag) against more-complex multi-stage retrieval systems (i.e., ReadAgent and RAPTOR) to draw conclusions about design priorities for next-generation RAG systems.
We believe that retrieval and long-context LMs are complementary in a variety of real-world applications.

\section{Conclusion}

This work examined whether complex multi-stage retrieval pipelines still justify their added complexity given the emergence of long-context LMs capable of processing tens of thousands of tokens.
Our controlled evaluation under systematically scaled token budgets shows that simpler methods like \dosrag can effectively match or even outperform multi-stage pipelines such as ReadAgent and RAPTOR in QA tasks, without intermediate summarization or agentic processing.

We identified four key strategies that contributed to \dosrag's performance:

1. Retrieving from \emph{original passages rather than generated summaries}, maintaining source fidelity and minimizing information loss.  

2. Prioritizing \emph{retrieval recall over precision}, ensuring critical information is included within the effective context window, even at the cost of some less relevant content.  

3. Reordering retrieved passages to \emph{preserve original passage order}, which proved particularly beneficial when dealing with large retrieval budgets.  

4. Favoring \emph{simple over complex} pipelines, while leveraging strong embedding and language models for robustness.

Based on these findings, we recommend establishing \dosrag as a simple yet strong baseline for future RAG evaluations, paired with state-of-the-art embedding and language models and benchmarked under matched token budgets, to ensure that any added complexity is justified by clear performance improvements as models continue to advance.

\section*{Limitations}
Although our results indicate that simpler retrieve-then-read approaches can match or outperform more intricate multi-stage RAG pipelines when paired with long-context language models, our study has several limitations that qualify the generality of these findings.

Our experiments focus on multiple-choice and short-answer reading comprehension tasks over single long documents. We used \gptFourO and \gptFourOmini as readers and Snowflake's Arctic-Embed as the embedding model for neural retrieval.
While these settings provide useful testbeds for long-context reasoning, it is unclear whether the trends hold for more diverse tasks such as open-ended generation, tasks that require reasoning over multiple documents, or complex reasoning that requires specialized domain knowledge (e.g., in scientific or legal domains). It also remains open whether the findings generalize to other reader LMs (proprietary or open-weight), or alternative retrieval setups with different embedding models. 
Future work should investigate whether the benefits of simply preserving document continuity extend to these settings, or whether specialized retrieval or summarization steps prove more valuable.

Efficiency is also a key factor in practice. While our comparisons matched token budgets for inference and show \dosrag competitive across both smaller and larger retrieval windows, we did not measure end-to-end costs of embedding and preprocessing. We estimate that more complex preprocessing, as in RAPTOR and ReadAgent, incurs additional cost, but future work should provide full cost analyses, especially for high-throughput or resource-limited scenarios.

\section*{Acknowledgments}
We would like to thank the anonymous reviewers for their helpful comments and feedback. 

\bibliography{custom}

\appendix

\section{Experimental Setup Details}\label{app:experimental-setup-details}

\subsection{Models and Computational Resources}
Throughout our experiments, we use the Snowflake Arctic-embed-m 1.5 model to embed queries and documents for retrieval, which has a size of 109M parameters \citep{merrick2024embeddingclusteringdataimprove}.

To better understand the effect of reader capability, we conduct experiments with \gptFourOmini (\texttt{"gpt-4o-mini-2024-07-18"}) and \gptFourO (\texttt{"gpt-4o-2024-11-20"}) as the reader language models. OpenAI does not publicly disclose the number of parameters for these models.

All experiments use greedy decoding for response generation. Our computational budget primarily consisted of API calls to OpenAI, with an estimated total token usage of 2 billion tokens (2B) across all experiments. Since inference was conducted via API, no local GPUs were used for model execution.

For retrieval and preprocessing, we used a local MacBook. The total compute time for retrieval and data preparation was approximately 12 CPU hours.

\subsection{Benchmark Licensing and Usage}

The benchmarks used in this study have the following license terms:
\begin{itemize}
    \item \inftybench: MIT License
    \item \quality: CC BY 4.0
    \item \narrativeqa: Apache-2.0 License
\end{itemize}
These datasets have been used strictly in accordance with their intended research purposes, as specified by their respective licenses. No modifications were made that would alter their intended scope or permitted usage. All evaluations conducted in this study fall within standard research practices, and no dataset derivatives have been deployed outside of a research context.

We did not conduct separate checks for personally identifiable information (PII) or offensive content beyond the dataset providers' original curation efforts. The responsibility for anonymization and content moderation lies with the original dataset creators. However, we relied on the fact that these benchmarks are widely used in research and released under established licenses, which include ethical considerations in their curation.

No personal data was stored, processed, or collected as part of this work. Additionally, no dataset derivatives were created, ensuring that any potential privacy risks remain within the scope of the original dataset publication.

\subsection{Hyperparameters}
In this study, we analyze the impact of retrieval hyperparameters on RAG performance. Unlike prior work, we do not train new models but instead evaluate how different retrieval depth, input token length, and chunking strategies influence final performance.

The primary hyperparameters studied include the maximum input length to the reader model. It varied from {500, 1K, 1.5K, 2K, 4K, 6K, 8K, 10K, 20K, 30K, 40K} tokens.

\subsection{Parameters for Packages}
For sentence segmentation, we use NLTK with its default model. For evaluation, we use the 'evaluate' package (\texttt{evaluate.load()}), computing the following metrics with default parameters:  
\begin{itemize}
    \item \textbf{F1-score}
    \item \textbf{BLEU-1, BLEU-4}
    \item \textbf{METEOR}
    \item \textbf{ROUGE-L}
\end{itemize}

All implementations are taken from the Hugging Face \texttt{evaluate} library, using the latest available version at the time of the experiments (\texttt{evaluate==0.4.3}). No modifications were made to the implementations.  

\subsection{Use of AI Assistance}
During this research, we used ChatGPT to assist with coding, debugging, and editing. Specifically:
\begin{itemize}
    \item Coding and Debugging: ChatGPT was used as a coding assistant for troubleshooting errors, generating boilerplate code, and refining scripts.
    \item Paper Writing and Editing: ChatGPT was used for grammar suggestions, phrasing improvements, and structural refinements of the paper. All technical content and research contributions were fully authored by the authors.
\end{itemize}
The final decisions on all implementations and manuscript edits were made by the authors.

\subsection{ReadAgent}

In our experiments, we adapt ReadAgent from its official public demo notebook with minimal changes.
Since many of the documents in our benchmarks do not contain reliable paragraph boundaries, we use individual sentences as the smallest unit for pre-processing and building ReadAgent's ``pages''.
Following \citet{lee2024humaninspiredreadingagentgist}, we allow ReadAgent to look up between 1 and 6 pages during inference (the best-performing range in the original paper).
In rare cases where the shortened pages plus gists still exceeded the token limit, we omitted those queries from evaluation (for instance, one document in \inftybench was dropped).

\subsection{RAPTOR}

We implement RAPTOR using the official repository.
To match our other systems, we use the NLTK library for sentence segmentation and the Snowflake Arctic-embed-m 1.5 embedding model \citep{merrick2024embeddingclusteringdataimprove} to embed and cluster passages.
In all experiments, we use \gptFourOmini to build the tree of hierarchical summaries to reduce API costs (though note that we experiment with both \gptFourOmini and \gptFourO as the downstream reader).

\subsection{Prompting}\label{app:prompting}

\begin{prompt}[title={Prompt \thetcbcounter: multiple-choice QA}, label=prompt:multiple_choice_qa]
[Start of Context]:

\{context\}

[End of Context]

[Start of Question]:

\{questionAndOptions\}

[End of Question]

[Instructions:]
Based on the context provided, select the most accurate answer to the question from the given options.
Start with a short explanation and then provide your answer as [[1]] or [[2]] or [[3]] or [[4]]. 
For example, if you think the most accurate answer is the first option, respond with [[1]].
\end{prompt}

\begin{prompt}[title={Prompt \thetcbcounter: QA generation}, label=prompt:qa_generation]
[Start of Context]:

\{context\}

[End of Context]

[Start of Question]:

\{question\}

[End of Question]

[Instructions:]
- Answer the question **only** based on the provided context.

- Keep the answer **short and factual** (preferably between 1-20 words).

- Do **not** provide explanations or additional details beyond what is necessary.

- If the answer is **not explicitly stated** in the context, respond with: "Not found in context."
\end{prompt}

\section{Comparing \gptFourOmini to \gptFourO}\label{app:gpt4o-vs-gpt4o-mini}

\begin{figure*}[t]
  \includegraphics[width=0.48\linewidth]{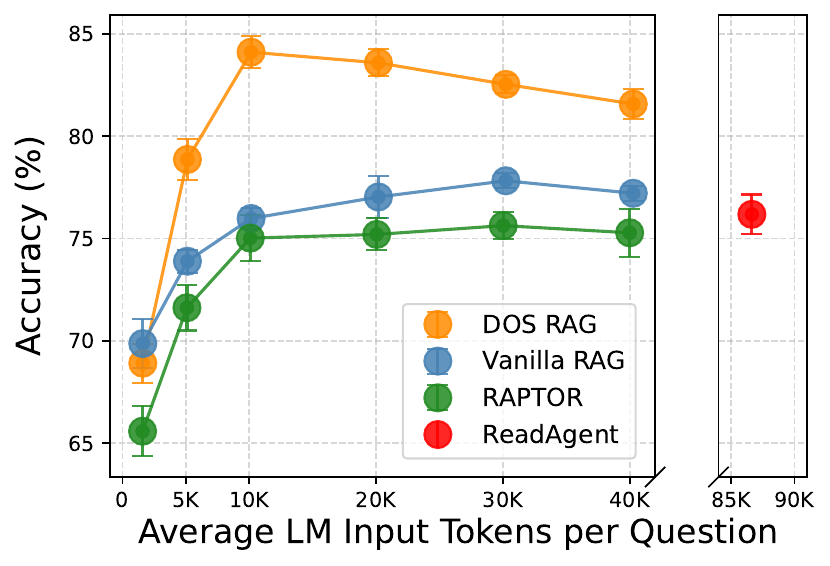} \hfill
  \includegraphics[width=0.48\linewidth]{images/infinity-bench-mc-4o.pdf}
  \caption { \inftybench En.MC performance of various multi-stage RAG systems and long-context baselines (mean  $\pm$ standard deviation over five runs). Comparison between \gptFourOmini (left) and \gptFourO (right) as the reader. \gptFourO generally achieves higher accuracy, with \dosrag peaking at a higher LM input token count, suggesting a larger effective context size. The ReadAgent results further indicate that \gptFourO can better utilize large context sizes, reaching performance levels generally comparable to the \dosrag results despite using an excessive number of input tokens.
  }
  \label{fig:infinity-bench-side-by-side}
\end{figure*}

Figure~\ref{fig:infinity-bench-side-by-side} provides a side-by-side comparison of \gptFourOmini and \gptFourO for the \inftybench results.

\section{Full Results}\label{app:full_results}

\paragraph{\inftybench Results.}
Table~\ref{tab:infinity-mc-gpt-mini} presents the \inftybench results for various systems and baselines using \gptFourOmini. Table~\ref{tab:infinity-mc-gpt-4o} reports the same results with \gptFourO. 

\begin{table*}[t]
\centering
\small
\begin{tabular}{lcccccc}
\toprule
 & \multicolumn{6}{c}{\textbf{Maximum Retrieval Token Budget}} \\
\cmidrule(lr){2-7}
\textbf{Method} & \textbf{1.5K} & \textbf{5K} & \textbf{10K} & \textbf{20K} & \textbf{30K} & \textbf{40K} \\
\midrule
\vanillarag & 
    69.9\% $\pm$ 1.2\% & 
    73.9\% $\pm$ 0.6\% & 
    76.0\% $\pm$ 0.5\% & 
    77.0\% $\pm$ 1.0\% & 
    77.8\% $\pm$ 0.4\% & 
    77.2\% $\pm$ 0.4\% \\
\dosrag & 
    68.9\% $\pm$ 1.0\% & 
    78.9\% $\pm$ 1.0\% & 
    \textbf{84.1}\% $\pm$ \textbf{0.8}\% & 
    83.6\% $\pm$ 0.7\% & 
    82.5\% $\pm$ 0.4\% & 
    81.6\% $\pm$ 0.7\% \\
\midrule
RAPTOR & 
    65.6\% $\pm$ 1.2\% & 
    71.6\% $\pm$ 1.1\% & 
    75.0\% $\pm$ 1.1\% & 
    75.2\% $\pm$ 0.8\% & 
    75.6\% $\pm$ 0.7\% & 
    75.3\% $\pm$ 1.2\% \\
\midrule
ReadAgent & \multicolumn{6}{c}{76.2\% $\pm$ 1.0\% \quad (\textit{Avg. Tokens:} 86K)} \\
\bottomrule
\end{tabular}
\caption{\inftybench En.MC performance of various systems with \gptFourOmini (mean $\pm$ standard deviation over five runs). 
ReadAgent uses its default configuration, and its average tokens-per-query is shown for comparison. \dosrag consistently outperforms all other methods for retrieval budgets of 5K tokens and above being the preferred choice in terms of both performance and efficiency.
}
\label{tab:infinity-mc-gpt-mini}
\end{table*}

\begin{table*}[t]
\centering
\small
\begin{tabular}{lcccccc}
\toprule
 & \multicolumn{6}{c}{\textbf{Maximum Retrieval Token Budget}} \\
\cmidrule(lr){2-7}
\textbf{Method} & \textbf{1.5K} & \textbf{5K} & \textbf{10K} & \textbf{20K} & \textbf{30K} & \textbf{40K} \\
\midrule
\vanillarag & 
    76.2\% $\pm$ 0.6\% & 
    82.6\% $\pm$ 0.8\% & 
    86.0\% $\pm$ 0.7\% & 
    86.9\% $\pm$ 0.5\% & 
    87.8\% $\pm$ 0.4\% & 
    86.6\% $\pm$ 0.4\% \\
\dosrag & 
    75.6\% $\pm$ 0.2\% & 
    85.9\% $\pm$ 0.6\% & 
    90.0\% $\pm$ 0.5\% & 
    91.4\% $\pm$ 0.2\% & 
    \textbf{93.1\% $\pm$ 0.5\%} & 
    91.9\% $\pm$ 0.7\% \\
\midrule
RAPTOR & 
    73.8\% $\pm$ 0.3\% & 
    79.0\% $\pm$ 0.4\% & 
    82.4\% $\pm$ 0.5\% & 
    84.4\% $\pm$ 0.6\% & 
    85.0\% $\pm$ 0.2\% & 
    85.9\% $\pm$ 0.4\% \\
\midrule
ReadAgent & \multicolumn{6}{c}{90.3\% $\pm$ 0.9\% \quad (\textit{Avg. Tokens:} 86K)} \\
\bottomrule
\end{tabular}
\caption{\inftybench En.MC performance of various systems with \gptFourO (mean $\pm$ standard deviation over five runs). 
ReadAgent uses its default configuration, and its average tokens-per-query is shown for comparison. \dosrag consistently outperforms all other methods for retrieval budgets of 5K tokens and above being the preferred choice in terms of both performance and efficiency.
}
\label{tab:infinity-mc-gpt-4o}
\end{table*}

\paragraph{\quality Results.}\label{app:quality_results}

\begin{figure}[t]
  \includegraphics[width=\columnwidth]{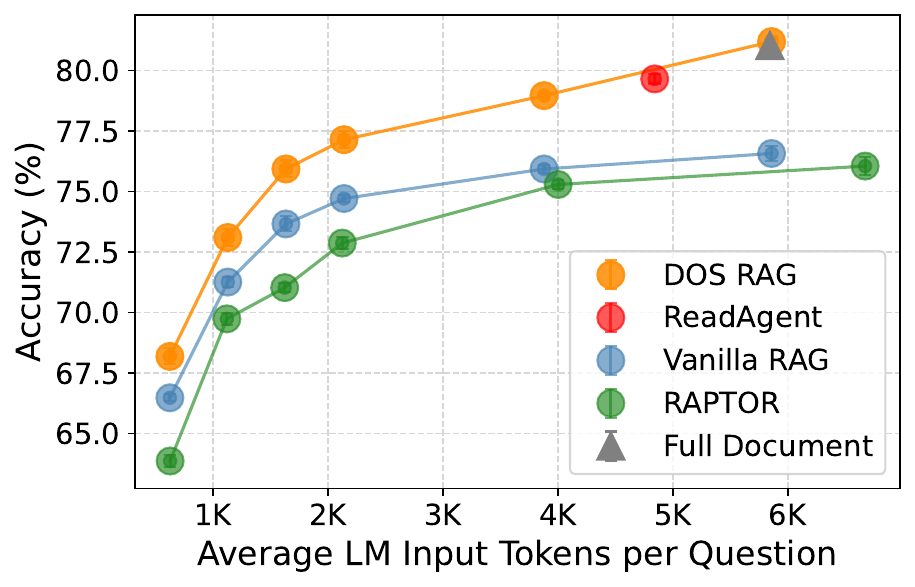}
  \caption{Accuracy progression with increasing LM input tokens for the
QuALITY development set with \gptFourOmini (mean $\pm$ standard
deviation over five runs)}
  \label{fig:quality-4o-mini-plot}
\end{figure}

Table~\ref{tab:quality-dev-gpt-mini} presents the \quality results for various systems and baselines using \gptFourOmini. Figure~\ref{fig:quality-4o-mini-plot} illustrates the accuracy progression as LM input tokens increase. Table~\ref{tab:quality-dev-gpt-4o} reports the same results but with \gptFourO as the reader.

\begin{table*}[t]
\centering
\small
\begin{tabular}{lcccccc}
\toprule
 & \multicolumn{6}{c}{\textbf{Maximum Retrieval Token Budget}} \\
\cmidrule(lr){2-7}
\textbf{Method} & \textbf{500} & \textbf{1K} & \textbf{1.5K} & \textbf{2K} & \textbf{4K} & \textbf{8K} \\
\midrule
\vanillarag & 
    66.5\% $\pm$ 0.2\% & 
    71.3\% $\pm$ 0.2\% & 
    73.7\% $\pm$ 0.3\% & 
    74.7\% $\pm$ 0.2\% & 
    75.9\% $\pm$ 0.2\% & 
    76.6\% $\pm$ 0.3\% \\
\dosrag & 
    68.2\% $\pm$ 0.3\% & 
    73.1\% $\pm$ 0.3\% & 
    75.9\% $\pm$ 0.4\% & 
    77.1\% $\pm$ 0.4\% & 
    79.0\% $\pm$ 0.1\% & 
    \textbf{81.2}\% $\pm$ \textbf{0.2}\% \\
\midrule
RAPTOR & 
    63.9\% $\pm$ 0.3\% & 
    69.7\% $\pm$ 0.3\% & 
    71.0\% $\pm$ 0.2\% & 
    72.9\% $\pm$ 0.3\% & 
    75.3\% $\pm$ 0.2\% & 
    76.3\% $\pm$ 0.4\% \\
\midrule
ReadAgent & \multicolumn{6}{c}{79.7\% $\pm$ 0.2\% \quad (\textit{Avg. Tokens:} 4.8K)} \\
\midrule
Full Document & \multicolumn{6}{c}{\textbf{81.0\% $\pm$ 0.3\%}
\quad (\textit{Avg. Tokens:} 5.8K)} \\
\bottomrule
\end{tabular}
\caption{QuALITY development set performance of various systems with \gptFourOmini (mean $\pm$ standard deviation over five runs). 
ReadAgent uses its default configuration, and its average tokens-per-query is shown for comparison. 
On QuALITY, prompting with entire documents gives the best accuracy. At 8K tokens, \dosrag effectively recovers the full document content and matches that performance; under tighter token budgets, \dosrag is the strongest method.
}
\label{tab:quality-dev-gpt-mini}
\end{table*}

\begin{table*}[t]
\centering
\small
\begin{tabular}{lcccccc}
\toprule
 & \multicolumn{6}{c}{\textbf{Maximum Retrieval Token Budget}} \\
\cmidrule(lr){2-7}
\textbf{Method} & \textbf{500} & \textbf{1K} & \textbf{1.5K} & \textbf{2K} & \textbf{4K} & \textbf{8K} \\
\midrule
\vanillarag & 
    73.0\% $\pm$ 0.2\% & 
    79.4\% $\pm$ 0.1\% & 
    82.0\% $\pm$ 0.1\% & 
    82.8\% $\pm$ 0.2\% & 
    85.0\% $\pm$ 0.2\% & 
    85.9\% $\pm$ 0.1\% \\
\dosrag & 
    74.1\% $\pm$ 0.3\% & 
    81.7\% $\pm$ 0.3\% & 
    84.8\% $\pm$ 0.1\% & 
    86.4\% $\pm$ 0.1\% & 
    88.9\% $\pm$ 0.2\% & 
    90.4\% $\pm$ 0.3\% \\
\midrule
RAPTOR & 
    71.8\% $\pm$ 0.2\% & 
    77.0\% $\pm$ 0.2\% & 
    79.6\% $\pm$ 0.3\% & 
    80.9\% $\pm$ 0.2\% & 
    83.7\% $\pm$ 0.2\% & 
    84.9\% $\pm$ 0.2\% \\
\midrule
ReadAgent & \multicolumn{6}{c}{87.4\% $\pm$ 0.3\% \quad (\textit{Avg. Tokens:} 4.2K)} \\
\midrule
Full Document & \multicolumn{6}{c}{\textbf{91.2\% $\pm$ 0.2\%}
\quad (\textit{Avg. Tokens:} 5.8K)} \\
\bottomrule
\end{tabular}
\caption{QuALITY development set performance of various systems with \gptFourO (mean $\pm$ standard deviation over five runs). ReadAgent uses its default configuration, and its average tokens-per-query is shown for comparison. 
On QuALITY, prompting with entire documents gives the best accuracy. Under token budgets, \dosrag is the strongest method.
}
\label{tab:quality-dev-gpt-4o}
\end{table*}

\paragraph{\narrativeqa Results.}\label{app:narrativeqa_results}
Table~\ref{tab:narrative-test-4o-mini} presents the results for the \narrativeqa test set across various systems and baselines, using \gptFourOmini as the reader. Some documents contain up to 404K tokens, far exceeding the 128K context size, which is why we do not report a full-document baseline. Due to issues with the original \narrativeqa download script, three out of 355 stories from the test set were inaccessible, as their document files were empty. Consequently, our results are reported for 352 documents and 10,391 questions for all methods.

\begin{table*}[t]
\centering
\small
\begin{tabular}{lcccccc}
\toprule
\textbf{Method} & \textbf{Token}  & \multicolumn{5}{c}{\textbf{Metric}} \\
\cmidrule(lr){3-7}
 & \textbf{Avg Spent / Budget} & \textbf{F1} & \textbf{BLEU-1} & \textbf{BLEU-4} & \textbf{ROUGE-L} & \textbf{METEOR} \\
\midrule
\vanillarag & 1.5K / 1.5K &
    15.1 &
    20.0 &
    3.7 &
    15.6 &
    21.3 \\
        & 5K / 5K &
    17.9 &
    21.1 &
    4.3 &
    18.4 &
    24.5 \\
        & 10K / 10K &
    18.8 &
    21.3 &
    4.4 &
    19.3 &
    25.7 \\
        & 19K / 20K &
    19.2 &
    21.4 &
    4.5 &
    19.8 &
    26.3 \\
        & 28K / 30K &
    19.4 &
    21.5 &
    4.5 &
    19.9 &
    26.6 \\
        & 35K / 40K &
    19.5 &
    21.6 &
    4.6 &
    19.9 &
    26.6 \\
\addlinespace[5pt]
\dosrag & 1.5K / 1.5K &
    16.3 &
    20.4 &
    3.9 &
    16.8 &
    22.5 \\
        & 5K / 5K &
    20.1 &
    21.7 &
    4.5 &
    20.6 &
    27.0 \\
        & 10K / 10K &
    21.2 &
    22.2 &
    4.8 &
    21.7 &
    28.5 \\
        & 19K / 20K &
    \textbf{22.1} &
    22.6 &
    5.0 &
    \textbf{22.5} &
    29.6 \\
        & 28K / 30K &
    \textbf{22.1} &
    \textbf{22.7} &
    5.0 &
    \textbf{22.5} &
     \textbf{29.8} \\
        & 35K / 40K &
    22.0 &
    \textbf{22.7} &
    \textbf{5.1} &
    22.3 &
    29.6 \\
    
\midrule
RAPTOR & 1.5K / 1.5K &
    10.5 &
    18.1 &
    2.9 &
    10.7 &
    16.4 \\
        & 5K / 5K &
    14.2 &
    19.6 &
    3.5 &
    14.5 &
    20.3 \\
        & 10K / 10K &
    15.9 &
    20.2 &
    3.9 &
    16.2 &
    22.3 \\
        & 19K / 20K &
    17.3 &
    20.8 &
    4.2 &
    17.6 &
    23.8 \\
        & 28K / 30K &
    17.8 &
    20.9 &
    4.3 &
    18.1 &
    24.5 \\
        & 37K / 40K &
    17.9 &
    21.1 &
    4.4 &
    18.2 &
    24.7 \\
    
\midrule
ReadAgent & 34K / --- &
    21.0 &
    22.2 &
    4.8 &
    21.4 &
    28.7 \\
\bottomrule
\end{tabular}
\caption{NarrativeQA test set performance of various systems and metrics with \gptFourOmini as the reader. For each method, the token budget and the actual average tokens used per question are shown; actual usage may fall below the budget when documents are shorter. \dosrag, with budgets of 20–40K tokens, outperforms all other methods across all metrics.}
\label{tab:narrative-test-4o-mini}
\end{table*}

\end{document}